\documentclass[11pt, a4paper, logo, copyright, nonumbering]{xiaomi}

\usepackage[numbers, sort&compress, square]{natbib}
\usepackage{dblfloatfix}
\usepackage[normalem]{ulem}
\usepackage{caption}
\usepackage{xspace}
\usepackage{pifont} 
\usepackage{multirow}
\usepackage{tcolorbox}
\usepackage{xltabular}
\usepackage{longtable}
\usepackage{hyperref}
\usepackage{wrapfig}
\usepackage{algorithm}
\usepackage{algorithmic}
\usepackage{amsfonts}
\usepackage{amsmath}
\usepackage{amssymb}
\usepackage{lineno}
\usepackage{multirow}
\usepackage{adjustbox}
\usepackage{pdfpages}

\usepackage[bottom]{footmisc}

\usepackage{CJKutf8}
\usepackage{setspace}
\usepackage{makecell}
\usepackage{graphicx}
\usepackage{subcaption}
\usepackage{multicol} 

\usepackage[utf8]{inputenc} 
\usepackage[T1]{fontenc}    
\usepackage{hyperref}
\usepackage{cleveref}
\usepackage{url}            
\usepackage{booktabs}       
\usepackage{amsfonts}       
\usepackage{nicefrac}       
\usepackage{microtype}      
\usepackage{xcolor}         
\usepackage{colortbl}
\usepackage{tcolorbox}
\usepackage{xspace}
\definecolor{BrickRed}{rgb}{.72,0,0}
\definecolor{darkgreen}{rgb}{0.0, 0.5, 0.0}
\definecolor{ForestGreen}{RGB}{34,139,34}
\definecolor{LakeBlue}{RGB}{0,61,153}
\definecolor{MiOrange}{RGB}{255,225,204}
\definecolor{Hex}{RGB}{225,213,231}

\hypersetup{
    colorlinks=true,
    allcolors=LakeBlue
}

\title{\centering GRNEdit: Efficient General Video Editing from a New Binary-Evidence Perspective in Generative Refinement Networks}

\titlerunning{Efficient General Video Editing from a New Binary-Evidence Perspective}

\author{
Feng Xie\textsuperscript{1,2,*},
Jiagao Hu\textsuperscript{2},
Fuhao Li\textsuperscript{2},
Zepeng Wang\textsuperscript{2},
Yuxuan Chen\textsuperscript{2},
Dahua Gao\textsuperscript{1,\textdagger},
Fei Wang\textsuperscript{2},
Daiguo Zhou\textsuperscript{2}
}

\institute{
\textsuperscript{1}Xidian University\\
\textsuperscript{2}MiLM Plus, Xiaomi Inc.
}

\begin{document}

\begingroup
\renewcommand{\thefootnote}{\fnsymbol{footnote}}
\footnotetext[1]{Work done during Feng Xie's internship at Xiaomi Inc.}
\footnotetext[2]{Corresponding author: Dahua Gao
(\texttt{dhgao@xidian.edu.cn}).}
\endgroup

\begin{abstract}
Instruction-based general video editing seeks to unify diverse editing operations within a single, intuitive interface. Existing approaches often rely on resource-intensive conditioning, using either heavyweight branches or costly source concatenation. Is there any efficient way to model editing intent? Thus, we introduce GRNEdit, a lightweight two-stage framework. GRN inspires our approach by encoding visual semantics through combinations of bits. Through task-specific fine-tuning, we take this representation further and recast editing semantics as local retain-or-flip decisions over individual bits. Source information is consequently modeled as coordinate-wise evidence supporting the observed binary states, while the GRN backbone remains responsible for resolving their global composition into coherent generative semantics. In Stage I, a compact encoder translates discrete source codes into continuous evidence signals, which GRN assimilates throughout binary refinement. Inspired by null-prompt training for classifier-free guidance, we further assign the null condition an editing-specific meaning: an empty instruction denotes no edit and is supervised through source reconstruction. This identity pathway not only implicitly strengthens evidence utilization and content preservation in Stage I, but also produces a source-preserving state in the same representation space as the edited state. Stage II can therefore directly compare each edited state with its source-preserving counterpart and use their discrepancy to revise unresolved target-bit decisions. Trained on only 0.6M pairs with less than 3\% conditioning parameters, GRNEdit-2B and GRNEdit-8B achieve scores of 4.03 and 4.18 on OpenVE-Bench. The 2B model outperforms multiple 14B open-source editors, while the 8B model performs on par with leading open-source editors.

Code resources are available at \url{https://github.com/Foxerity/GRNEdit}.
\end{abstract}

\maketitle

\clearpage

\begin{figure}[!t]
    \centering
    \includegraphics[
    width=\linewidth,
    height=0.60\textheight,
    keepaspectratio
]{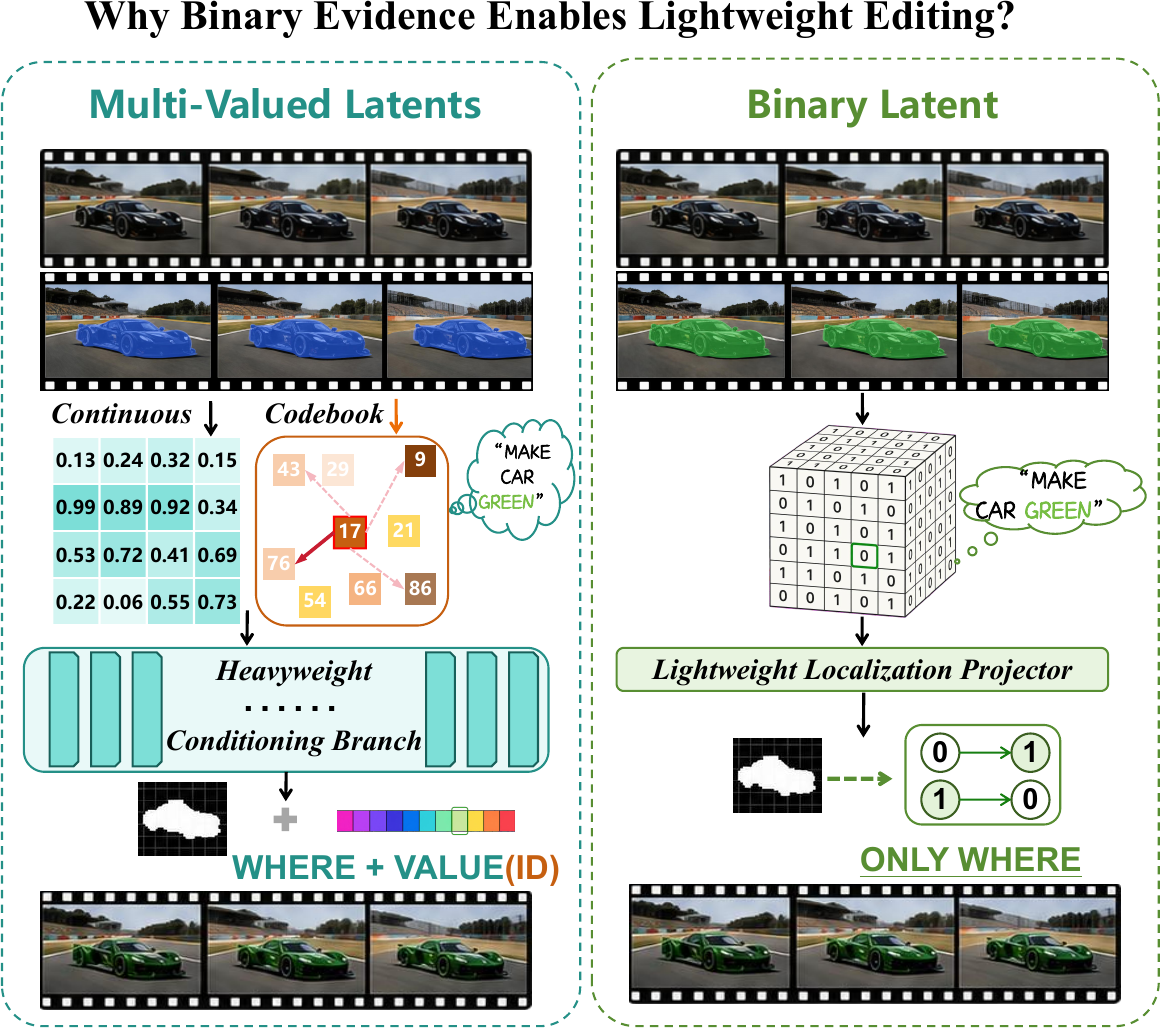}
    \caption{Continuous and codebook latents require predicting both edit locations
    and target values/indices, whereas each binary bit has a unique alternative
    state, enabling lightweight localization-only control.}
    \label{fig:binary_evidence_motivation}
\end{figure}

\section{Introduction}
Instruction-based general video editing unifies diverse editing operations under open-ended instructions \cite{insv2v2024,insvie2025,ditto2026,openve2025,ivebench2026,viva2026,cotedit2026}. Most existing systems operate in the continuous spaces of large diffusion or flow backbones. Their source-conditioning mechanism must both interpret editing intent and construct representations compatible with the backbone's generative states. This typically requires heavyweight branches or costly source concatenation. As generative backbones scale, the conditioning cost and optimization burden grow accordingly.

Such conditioning is commonly implemented with copied blocks, reference streams, adapters, or source tokens processed jointly by the backbone \cite{controlnet2023,vace2025,videodirector2025,easyv2v2026}. Despite their different forms, these mechanisms must simultaneously localize the intended change and produce dense, target-aligned corrections. Increasing their capacity improves representation alignment but raises training and inference costs, while shrinking them makes high-dimensional corrections harder to learn. We therefore ask: Can content synthesis remain with the backbone, while a lightweight branch models only source editing evidence?

The Generative Refinement Network (GRN) provides a generative interface distinct from diffusion and autoregressive models \cite{grn2026}. Its Hierarchical Binary Quantization (HBQ) encodes video latents as hierarchical binary codes, which GRN predicts by repeatedly refining the full bit map. This replaces diffusion-style continuous regression with explicit binary decisions. Unlike causal autoregression, it also avoids a long chain of irreversible token predictions.

When the source and target are quantized by the HBQ tokenizer, they share aligned binary coordinates. At each coordinate, the source provides an observed state, and editing requires deciding whether to retain or flip it. Compared with predicting continuous feature corrections or selecting among multiple codebook entries, this binary choice substantially reduces the local search burden on the conditioning path, as illustrated in Fig.~\ref{fig:binary_evidence_motivation}. This observation motivates our \emph{binary evidence perspective}: the source is treated not as a residual target representation, but as evidence for target-bit decisions. Continuous embeddings carry this evidence, while the GRN backbone resolves global bit composition.

Stage~I realizes this separation through lightweight source encoders that map source bits into hidden evidence messages at selected Transformer chunks. Instruction-aware modulation adjusts their influence, while GRN's native target-bit likelihood determines whether each source state should be retained or overturned. The conditioning path therefore learns the relevance of source observations, leaving target synthesis to the pretrained backbone.

To anchor evidence injection in preservation semantics, we draw inspiration from null-prompt training in classifier-free guidance ~\cite{cfg2022} and reinterpret conventional condition dropout as an identity-aligned null condition. Instead of asking the model to fit the edited target after the text condition has been removed, we define an empty instruction as no edit and supervise it through source reconstruction. The contrast between editing and no editing strengthens Stage I's interpretation of editing semantics and use of source evidence, while improving content preservation. It also enables the same editor to produce a source-preserving state in the same parameterization as the edited state, providing a natural reference for subsequent revision.

Using this reference, Stage II freezes Stage I and employs a lightweight Bit-Margin Router to compare edited and source-preserving states and revise unresolved target-bit decisions in logit space. The revised logits are optimized with GRN's native target-bit likelihood objective. Source evidence first participates in generation decisions in Stage I; the identity condition then establishes a preservation reference; and Stage II turns that reference into targeted revisions of residual errors. Throughout the process, target content synthesis remains the responsibility of the GRN backbone.

Our contributions are summarized as follows:

\begin{itemize}
    \item We introduce a new binary evidence perspective for general video editing, enabling efficient separation of generative capability from editing control and unified modeling of diverse editing intents.
    \item We introduce GRNEdit, a lightweight two-stage framework that rivals leading video editing models with minimal conditioning parameters and training data.
    \item We redesign classifier-free null conditioning for editing by anchoring the empty instruction to source reconstruction. This identity anchor sharpens edit/no-edit semantics, improves source consistency, and bridges the two stages.
\end{itemize}

\begin{figure*}[t]
    \centering
    \includegraphics[width=\textwidth]{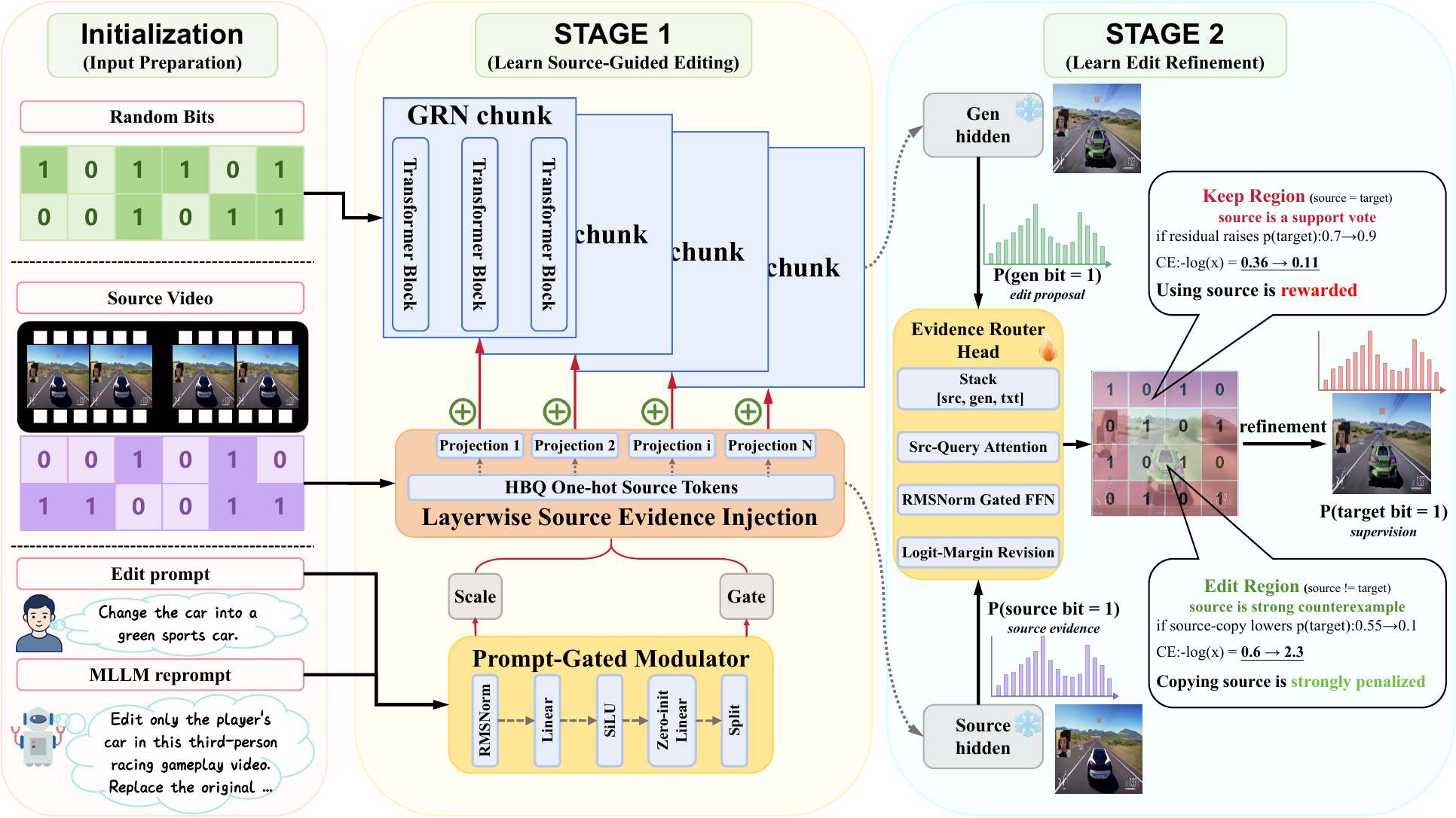}
    \caption{Overview of GRNEdit.
    Given a source video and an editing instruction, Stage I injects
prompt-modulated source evidence into the GRN chunks through lightweight
per-chunk projectors, each containing only about 3 M parameters for
GRNEdit-2B and 7 M for GRNEdit-8B. Stage II refines the binary
predictions by treating source evidence as support in keep regions and
as counter-evidence in edit regions.}
    \label{fig:pipeline}
\end{figure*}

\section{Related Work}
\subsection{Video Generation Paradigms}
Video generation spans continuous iterative models and discrete autoregressive models \cite{salt2026,resamplingforcing2026,unitemp2026,starflowv2026}. Diffusion systems operate on continuous latents through denoising or vector-field integration \cite{ddpm2020,flowmatching2023,hunyuanvideo2024,wan2025}. Visual autoregressive models optimize discrete likelihoods but typically commit to token- or scale-wise generation orders \cite{videopoet2024,var2024,infinity2025}. GRN combines discrete HBQ prediction with global random refinement, repeatedly revisiting the full binary state rather than following a fixed causal order \cite{grn2026}. This globally revisable generator forms the backbone of GRNEdit.

\subsection{From Continuous to Binary Representations}
Diffusion and flow models operate on continuous VAE latents \cite{ldm2022,hunyuanvideo2024,wan2025}, whereas VQ-based models predict discrete codebook indices \cite{vqvae2017,videopoet2024,var2024}; bitwise representations replace this multi-way prediction with binary decisions. Infinity introduces bitwise autoregression, while GRN extends hierarchical binary quantization to images and videos \cite{infinity2025,grn2026,infinitystar2026}. Despite higher compression, GRN matches or surpasses continuous VAEs in reconstruction, reporting 0.56 vs.\ 0.87 rFID on ImageNet and near-parity on video.

\subsection{Instruction-Based General Video Editing}
Instruction-based video editing now covers object insertion, removal, and replacement, background modification and stylization. InsV2V, InsViE-1M, Ditto, and OpenVE-3M expand paired training data, while OpenVE-Bench and IVEBench evaluate instruction following, source preservation, and temporal quality \cite{insv2v2024,insvie2025,ditto2026,openve2025,ivebench2026}. VideoDirector and InstructVEdit study conditional control and architecture-data co-design; VEGGIE, VIVA, and CoT-Edit introduce multimodal reasoning, grounding, planning, or reward optimization \cite{videodirector2025,instructvedit2025,veggie2025,viva2026,cotedit2026}.

\subsection{Source Conditioning and Parameter-Efficient Adaptation}
Source conditioning follows several designs. InstructPix2Pix and InsViE fuse source and noisy latents before a common projection; EasyV2V concatenates separately embedded source tokens, while Omni-Video adapts multimodal clues into diffusion conditioning \cite{instructpix2pix2023,insvie2025,easyv2v2026,omnivideo2025}. ICVE uses weight-shared source/edit streams, whereas ControlNet and VACE add explicit branches or adapters \cite{icve2025,controlnet2023,vace2025}. Shared weights and LoRA reduce trainable parameters \cite{lora2022,adalora2023}, yet these methods still express control through continuous backbone states. GRNEdit instead models evidence over binary decisions.

\section{Method}
\label{sec:method}

\subsection{Binary Evidence Formulation}
\label{sec:binary_evidence}

Given a source video $V^s$ and an open-ended instruction $c$, general video editing generates $\widehat V^e=F_{\Theta}(V^s,c)$ while preserving unrelated appearance and motion. Using GRN's binary interface, we cast this task as source-referenced target-bit prediction. HBQ represents videos as hierarchical bit maps, which GRN predicts by iteratively refining all coordinates in parallel. The frozen video encoder $\mathcal{E}$ and HBQ tokenizer $\mathcal{Q}_M$ produce $Y^u=\mathcal{Q}_M(\mathcal{E}(V^u))\in\{0,1\}^{N\times D}$ for $u\in\{s,e\}$, where $N$ and $D$ denote visual positions and binary channels. The edit map $A^e=Y^s\oplus Y^e$ marks whether each source bit is retained ($A^e_{n,d}=0$) or flipped ($A^e_{n,d}=1$). For fixed $Y^s$, this is an exact reparameterization of $Y^e$: it preserves the global code space while anchoring each target-bit decision to the source. This source-relative formulation matches GRN's native objective. For two-class logits $Z_{n,d}\in\mathbb{R}^2$, define the source-aligned margin
\begin{equation}
\kappa_{n,d}
=(2Y^s_{n,d}-1)(Z_{n,d,1}-Z_{n,d,0}).
\label{eq:source_margin}
\end{equation}
Then $\sigma(\kappa_{n,d})$ is the retain probability, and the target-bit cross-entropy becomes
\begin{equation}
\begin{aligned}
\ell_{n,d}={}&-(1-A^e_{n,d})\log\sigma(\kappa_{n,d})\\
&-A^e_{n,d}\log\sigma(-\kappa_{n,d}),
\end{aligned}
\label{eq:relative_ce}
\end{equation}
 expressing target prediction as retain-or-flip supervision.

Let $\mathcal{M}^s$ collect the continuous source messages introduced below. We call them \emph{source evidence} because their effect is measured by the change in $\kappa_{n,d}$ relative to $\mathcal{M}^s=\mathbf{0}$, without a separate feature target. Supervision comes only from the final bit likelihood; thus, the conditioning path learns how source observations bias target decisions, while GRN remains responsible for global content synthesis.

GRN's global random refinement keeps this evidence revisable. At progress $p$, it forms
\begin{equation}
X_p(Y)=S_p\odot Y+(1-S_p)\odot Y^{\mathrm{rand}},
\label{eq:grn_refinement}
\end{equation}
where $S_p$ is a schedule-controlled binary selection mask and $Y^{\mathrm{rand}}$ contains random bits. GRN embeds $\operatorname{one-hot}(X_p)$ with $W_x$, processes it with the text condition $C_{\mathrm{main}}$ and progress embedding $e_p$ through the Transformer $G_{\theta}$, and applies the bit classifier $B_{\phi}$ to states selected by $\operatorname{Sel}_{\mathrm{vis}}$. Because GRN revisits every bit at each step, source evidence can be repeatedly reinterpreted rather than irreversibly committed.

\subsection{GRNEdit: A Two-Stage Evidence Editing Framework}
\label{sec:grnedit_overview}

As illustrated in Fig.~\ref{fig:pipeline}, GRNEdit organizes source-evidence injection, identity anchoring, and identity-referenced bit-margin revision into the two-stage pipeline summarized in Algorithm~\ref{alg:grnedit}. Where \(\operatorname{oh}(\cdot)\) denotes bitwise one-hot encoding.

\begin{algorithm}[t]
\caption{Two-stage training pipeline of GRNEdit}
\label{alg:grnedit}
\begin{algorithmic}[1]
\REQUIRE Source and target bits $(Y^s,Y^e)$; conditions
$(C_{\mathrm{main}},C_{\mathrm{edit}})$; identity probability $\rho$
\ENSURE Stage-I editor $(\bar{\theta},\bar{\eta},\bar{\phi})$ and
Bit-Margin Router $\psi$

\STATE \textbf{Stage I: Evidence assimilation and identity anchoring}
\FOR{each training sample}
    \STATE Sample refinement progress $p$ and $b\sim\mathrm{Bernoulli}(\rho)$
    \IF{$b=1$}
        \STATE $(Y^\star,C_{\mathrm{main}}^\star,C_{\mathrm{edit}}^\star)
        \gets(Y^s,C_{\varnothing},C_{\varnothing})$
        \COMMENT{identity}
    \ELSE
        \STATE $(Y^\star,C_{\mathrm{main}}^\star,C_{\mathrm{edit}}^\star)
        \gets(Y^e,C_{\mathrm{main}},C_{\mathrm{edit}})$
        \COMMENT{edit}
    \ENDIF
    \STATE $X_p^\star\gets X_p(Y^\star)$
    \STATE $Z^\star\gets
    \textsc{StageI}(X_p^\star,Y^s,
    C_{\mathrm{main}}^\star,C_{\mathrm{edit}}^\star,p)$
    \STATE Update $(\theta,\eta,\phi)$ using
    $\mathcal{L}_{\mathrm{bit}}(Z^\star,Y^\star)$
\ENDFOR

\STATE Freeze $(\theta,\eta,\phi)\rightarrow
(\bar{\theta},\bar{\eta},\bar{\phi})$

\STATE \textbf{Stage II: Identity-referenced bit-margin revision}
\FOR{each training sample}
    \STATE $X_p^g\gets X_p(Y^e)$
    \STATE $\mathcal{T}^g\gets
    (X_p^g,Y^s,C_{\mathrm{main}},C_{\mathrm{edit}},p)$
    \STATE $\mathcal{T}^r\gets
    (Y^s,Y^s,C_{\varnothing},C_{\varnothing},1)$
    \STATE $(H^g,H^r)\gets
    \textsc{FrozenStageIHidden}(\mathcal{T}^g,\mathcal{T}^r)$
    \STATE $(Z^g,Z^r)\gets
    (B_{\bar{\phi}}(H^g),B_{\bar{\phi}}(H^r))$
    \STATE $R\gets\mathcal{R}_{\psi}
    (\operatorname{sg}(H^g),\operatorname{sg}(H^r),C_{\mathrm{edit}})$
    \STATE $Z^{\mathrm{rev}}\gets
    \textsc{ReviseMargin}(Z^g,Z^r,R)$
    \STATE Update $\psi$ using
    $\mathcal{L}_{\mathrm{bit}}(Z^{\mathrm{rev}},Y^e)$
\ENDFOR
\end{algorithmic}
\end{algorithm}

\paragraph{Identity-Aligned Null Condition}
\label{sec:identity}

Classifier-free guidance (CFG) contrasts a conditional prediction with a null branch learned through condition dropout, but this branch lacks editing-specific preservation semantics. For editing, the null operation is identity: an empty instruction denotes no edit. We therefore reinterpret condition dropout as identity supervision. For $b\sim\operatorname{Bernoulli}(\rho)$, one draw jointly switches conditions, target, and perturbed state:
\begin{equation}
\begin{aligned}
(Y^\star,C_{\mathrm{main}}^\star,C_{\mathrm{edit}}^\star)
&=
\begin{cases}
(Y^e,C_{\mathrm{main}},C_{\mathrm{edit}}), & b=0,\\
(Y^s,C_{\varnothing},C_{\varnothing}), & b=1,
\end{cases}\\
X_p^\star
&=X_p(Y^\star).
\end{aligned}
\label{eq:identity_targets}
\end{equation}
The $C_{\mathrm{edit}}$ predicts $Y^e$ under the instruction; the identity path reconstructs $Y^s$ from source states under null conditions.

With $Z_j^\star$ denoting Stage~I logits and $\omega_j$ normalized span weights, both paths optimize $\mathcal{L}_{\mathrm{I}}=\sum_j\omega_j\ell_{\mathrm{bit}}(Z_j^\star,Y_j^\star)$ using $\ell_{\mathrm{bit}}$ from Eq.~\eqref{eq:relative_ce}. Sharing parameters, HBQ coordinates, and hidden spaces turns this contrast into a low-cost signal for learning when source evidence should be preserved or overridden, strengthening Stage~I evidence utilization and content preservation. Identity training across progress yields Stage~II's reference. At its clean endpoint $p=1$, $Y^s$ serves as both the current state and source evidence. The frozen Stage~I editor thus provides a coordinate-aligned, source-preserving hidden state for residual revision, without a separate encoder.

\begin{figure*}[t]
    \centering
    \includegraphics[width=\textwidth]{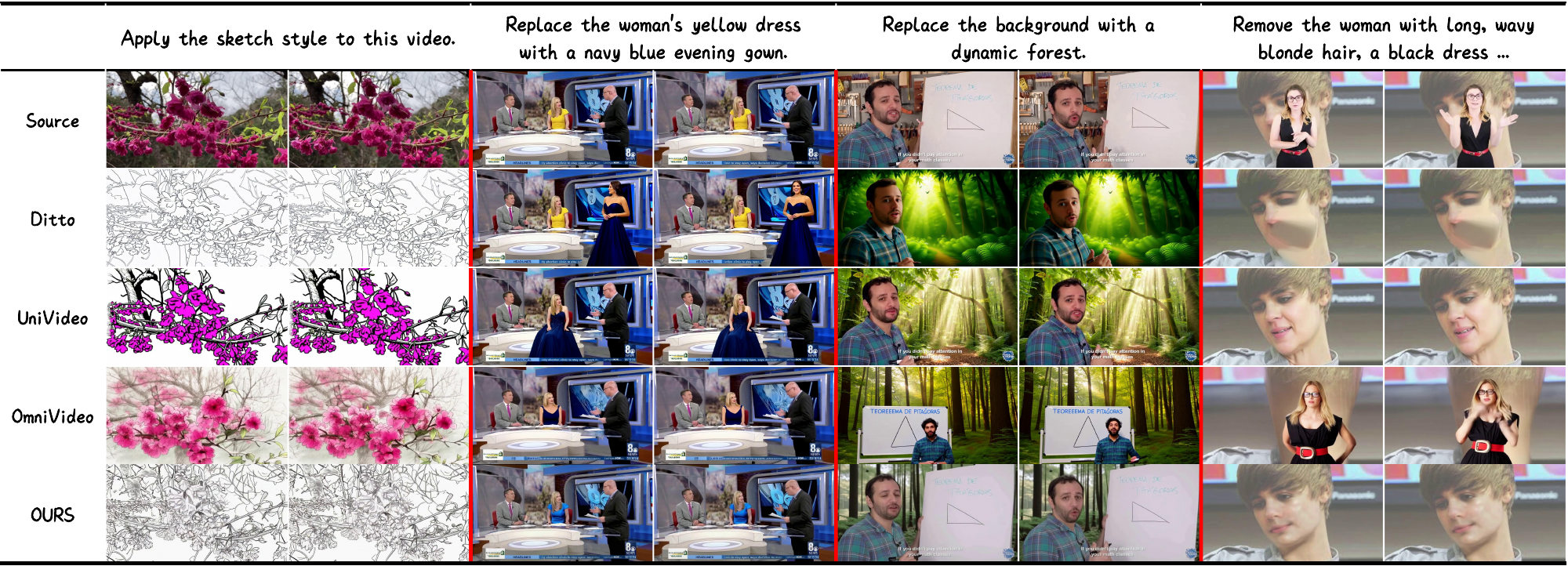}
    \caption{Qualitative comparison across diverse editing tasks.
    GRNEdit performs more accurate edits while better preserving unedited content, whereas competing methods often miss the intended change or disrupt scene consistency.}
    \label{fig:qualitative}
\end{figure*}

\paragraph{Stage I: Layerwise Source Evidence Injection}
\label{sec:stage1}

With identity defining what remains unchanged, Stage~I learns how source evidence enters editing. Source and target share HBQ coordinates, but successive GRN chunks use different hidden bases; we therefore adapt source bits at the input and selected boundaries. Let $q^s=\operatorname{oh}(Y^s)\in\{0,1\}^{N\times 2D}$. For chunk $k$,
\begin{equation}
\begin{aligned}
u_k
&=\operatorname{RMSNorm}\!\left(
    \operatorname{sg}([\bar{C}_{\mathrm{edit}};e_p])
\right),\\
[a_k;g_k]
&=W_k^{\mathrm{out}}
  \operatorname{SiLU}(W_k^{\mathrm{in}}u_k),\\
M_k^s
&=\mathcal{A}_k(q^s)\odot g_k\odot(1+a_k).
\end{aligned}
\label{eq:evidence_module}
\end{equation}
Here $\mathcal{A}_k$ maps source bits into the $C$-dimensional basis of chunk $k$, while the pooled edit condition $\bar{C}_{\mathrm{edit}}$ and progress embedding $e_p$ produce the gate $g_k\in\mathbb{R}^C$ and residual scale $a_k\in\mathbb{R}^C$. Thus, $M_k^s\in\mathbb{R}^{N\times C}$ is instruction-aware source evidence. Let $H_k$ be the packed state, $m_k\in\{0,1\}$ select boundaries, and $\Pi_{\mathrm{vis}}$ restrict additions to visual spans. For the $K$ Transformer chunks indexed by $k=0,\ldots,K-1$,
\begin{equation}
\begin{aligned}
H_0
&=[W_x\operatorname{oh}(X_p);
   C_{\mathrm{main}};e_p],\\
H_{k+1}
&=\operatorname{Chunk}_k\!\left(
    H_k+m_k\Pi_{\mathrm{vis}}(M_k^s)
\right).
\end{aligned}
\label{eq:layerwise_injection}
\end{equation}
This preserves GRN's native packed sequence, full attention, and bit head. Because the messages are supervised only through edited target-bit likelihood, the conditioning path learns how source observations should bias each decision, while GRN retains responsibility for target synthesis.

\paragraph{Stage II: Identity-Referenced Bit-Margin Revision}
\label{sec:stage2}

The identity path turns the CFG-inspired contrast into editing-aligned revision. Stage~II freezes Stage~I and reuses its clean endpoint as the preservation reference, rather than an unconditional or negative state. Let $\bar F$ and $B_{\bar\phi}$ denote the frozen visual-hidden mapping and bit head, and abbreviate $\mathcal C=(C_{\mathrm{main}},C_{\mathrm{edit}})$ and $\mathcal C_{\varnothing}=(C_{\varnothing},C_{\varnothing})$:
\begin{equation}
\begin{aligned}
H^g
&=\bar F(X_p^g;Y^s,\mathcal C,p),\\
H^r
&=\bar F(Y^s;Y^s,\mathcal C_{\varnothing},1).
\end{aligned}
\label{eq:aligned_states}
\end{equation}
Their logits are $Z^g=B_{\bar\phi}(H^g)$ and $Z^r=B_{\bar\phi}(H^r)$. Sharing the frozen editor keeps both states coordinate-aligned.

At each coordinate, the Bit-Margin Router queries with $H^r$ and attends to the aligned reference, generation, and pooled condition as three key--value tokens. Q/K RMS normalization and a GRN-style gated FFN are followed by a zero-initialized per-bit projection. Let $\widetilde H^u=\operatorname{sg}(H^u)$ for $u\in\{g,r\}$. The router predicts a $\tanh$-bounded $R\in(-1,1)^{N\times D}$ and applies it to the aligned margin discrepancy:
\begin{equation}
\begin{aligned}
\delta m
&=m(Z^r)-\operatorname{sg}\!\left(m(Z^g)\right),\\
R
&=\mathcal R_\psi(
    \widetilde H^g,\widetilde H^r,C_{\mathrm{edit}}
),\\
Z^{\mathrm{rev}}
&=Z^g+\tfrac12
  (R\odot\delta m)\otimes(-1,1).
\end{aligned}
\label{eq:stage2_revision}
\end{equation}
Here $m(Z)=Z_{\cdot,\cdot,1}-Z_{\cdot,\cdot,0}$. The symmetric update gives $m(Z^{\mathrm{rev}})=m(Z^g)+R\odot\delta m$ while preserving the mean logit. The router therefore learns only a per-bit signed coefficient on this discrepancy, rather than synthesizing a new correction direction. Hence, $R=0$ recovers Stage~I, $R\to1$ approaches the identity-reference margin, and $R<0$ moves away when required by the edit. Zero initialization therefore leaves Stage~I unchanged before training. Overall, Stage~I injects source evidence during synthesis; identity anchors preservation; and Stage~II revises residual bit decisions against that anchor under the edit condition.

\begin{table}[t]
\centering
\resizebox{\linewidth}{!}{
\begin{tabular}{@{}ll@{\hspace{3pt}/\hspace{3pt}}lcccccccc@{}}
\toprule
Method & \multicolumn{2}{c}{Params / Time} &
Overall & G.Style & Bkg.Chg. & L.Chg. & L.Rem. & L.Add. & Sub. & Cond. Params \\
\midrule
Runway Aleph (Commercial)
& \multicolumn{2}{l}{Commercial}
& 4.49 & 4.41 & 4.40 & 4.31 & 4.64 & 4.36 & 4.21 & N/A \\
\midrule
VACE (ICCV 2025)
& 14B & 545s
& 3.01 & 3.46 & 2.81 & 2.47 & 3.99 & 1.76 & 4.41 & 3.05B \\

Omni-Video (Tech. Rep. 2025)
& 11B & 312s
& 3.66 & 3.41 & \underline{4.11} & 3.75 & 4.52 & 2.80 & \textbf{4.95} & $>4.6$B \\

InsViE (ICCV 2025)
& 2B & 64s
& 3.25 & 3.63 & 2.68 & 2.82 & 3.56 & 2.25 & 4.77 & $\sim$59M \\

Lucy-Edit (Tech. Rep. 2025)
& 5B & 36s
& 3.77 & 3.64 & 3.25 & 3.93 & 3.95 & 3.92 & 4.23 & $\sim$0.6M \\

Kiwi-Edit (arXiv 2026)
& 8B & 45s
& \underline{4.17} & 4.24 & 4.01 & \underline{4.30} & \underline{4.60} & 4.13 & 4.23 & 3B \\

DITTO (CVPR 2026)
& 14B & 611s
& 3.44 & \textbf{4.48} & 3.52 & 2.89 & 3.53 & 2.48 & 3.69 & 3.17B \\

OpenVE-Edit (arXiv 2025)
& 5B & 150s
& 3.89 & 4.24 & 4.10 & 3.80 & 3.50 & 3.41 & 3.98 & $>3$B \\

LoomVideo (arXiv 2026)
& 5+8B & 166s
& 4.09 & \underline{4.44} & 3.91 & 4.02 & 4.22 & 4.21 & 4.66 & 89M \\

UniVideo (ICLR 2026)
& 14B & 893s
& \textbf{4.18} & 4.05 & 3.94 & \textbf{4.33} & 4.42 & \textbf{4.41} & 4.56 & $>7$B \\

Lance (Tech. Rep. 2026)
& 7.1B & 99s
& 4.01 & 3.99 & 4.01 & 3.72 & 4.17 & \underline{4.29} & 4.17 & N/A \\

\midrule
GRNEdit-2B (Ours)
& 2B & 39s
& 4.03 & 4.36 & 3.80 & 3.93 & 4.56 & 3.84 & 4.72 & 37M \\

GRNEdit-8B (Ours)
& 8B & 84s
& \textbf{4.18} & 4.37 & \textbf{4.12} & 4.09
& \textbf{4.73} & 3.86 & \underline{4.86} & 45M \\
\bottomrule
\end{tabular}
}
\caption{Comparison on OpenVE-Bench. Higher is better.  Cond. Params counts structural parameters beyond the listed generator.}
\label{tab:openve_main}
\end{table}

\begin{table}[!htbp]
\centering

\begingroup
\setlength{\tabcolsep}{2.2pt}
\renewcommand{\arraystretch}{1.05}

\resizebox{\linewidth}{!}{%
\begin{tabular}{@{}l*{16}{c}@{}}
\toprule
& \multicolumn{4}{c}{Add}
& \multicolumn{4}{c}{Remove}
& \multicolumn{4}{c}{Replace}
& \multicolumn{4}{c}{Style} \\
\cmidrule(lr){2-5}
\cmidrule(lr){6-9}
\cmidrule(lr){10-13}
\cmidrule(l){14-17}

Method
& $S_{\mathrm{EA}}$ & $S_{\mathrm{VN}}$ & $S_{\mathrm{VQ}}$ & $S$
& $S_{\mathrm{EA}}$ & $S_{\mathrm{VN}}$ & $S_{\mathrm{VQ}}$ & $S$
& $S_{\mathrm{EA}}$ & $S_{\mathrm{VN}}$ & $S_{\mathrm{VQ}}$ & $S$
& $S_{\mathrm{EA}}$ & $S_{\mathrm{VN}}$ & $S_{\mathrm{VQ}}$ & $S$ \\
\midrule

InsViE
& 2.60 & 3.10 & 3.46 & 3.05
& 2.44 & 3.76 & 3.29 & 3.16
& 2.10 & 3.91 & 3.49 & 3.17
& 8.17 & 8.21 & 7.35 & 7.91 \\

Lucy-Edit
& 6.47 & 5.70 & 6.77 & 6.31
& 7.02 & 6.88 & 6.81 & 6.90
& 7.08 & 6.21 & 6.88 & 6.72
& 4.65 & 4.67 & 5.17 & 4.83 \\

DITTO
& 6.70 & \textbf{7.57} & \underline{8.41} & \underline{7.56}
& 5.48 & 6.67 & \underline{6.93} & 6.36
& 4.56 & 7.21 & 7.96 & 6.58
& \underline{9.20} & 9.07 & 8.77 & 9.01 \\

ReCo
& \textbf{8.54} & \underline{7.55} & \textbf{8.61} & \textbf{8.23}
& \underline{7.28} & \underline{6.90} & 6.82 & \underline{7.00}
& \textbf{9.43} & \textbf{8.01} & \textbf{8.77} & \textbf{8.74}
& \textbf{9.42} & \underline{9.19} & \underline{8.90} & \textbf{9.17} \\

OURS-2B
& \underline{7.53} & 6.91 & 7.44 & 7.29
& \textbf{8.39} & \textbf{7.76} & \textbf{7.81} & \textbf{7.99}
& \underline{8.60} & \underline{7.24} & \underline{7.99} & \underline{7.94}
& 9.03 & \textbf{9.23} & \textbf{8.97} & \underline{9.08} \\

\bottomrule
\end{tabular}%
}

\endgroup

\caption{Comparison on ReCo-Bench across four editing tasks.
Results are ranked within each task; higher is better.}
\label{tab:reco_main}
\end{table}

\section{Experiments}
\label{sec:experiments}

\subsection{Experimental Setup}
\label{sec:experimental_setup}

\paragraph{Evaluation.}
We evaluate GRNEdit on OpenVE-Bench and ReCo-Bench~\cite{reco2026}. OpenVE-Bench covers eight editing tasks and reports category-wise and Overall scores. ReCo-Bench evaluates Add, Replace, Remove, and Style using $S_{\mathrm{EA}}$, $S_{\mathrm{VN}}$, $S_{\mathrm{VQ}}$, and the aggregate score $S$. Higher is better for all metrics. Open-source baselines \cite{runwayaleph2025,omnivideo2025,lucyedit2025,kiwiedit2026,univideo2026,lance2026,LoomVideo2026} in the tables follow their public settings and native resolutions; Runway Aleph is included only as a commercial reference.

\paragraph{Implementation details.}
We construct a task-balanced set of 0.6M training pairs from the public OpenVE-3M dataset. For each sample, we use the Qwen3-VL-Flash API to generate a source-aware reprompt from the source video and raw instruction. The reprompt is concatenated with the original instruction. The models are trained for 60K optimization steps using AdamW with $\beta_1=0.9$ and $\beta_2=0.999$, while the learning rate decays from $4\times10^{-5}$ to $1\times10^{-5}$. Training uses 16  H200 GPUs. Further details on implementation, sampling, and VLM prompting are provided in Appendix\,A1.

\begin{figure}[t]
    \centering
    \includegraphics[width=\columnwidth]{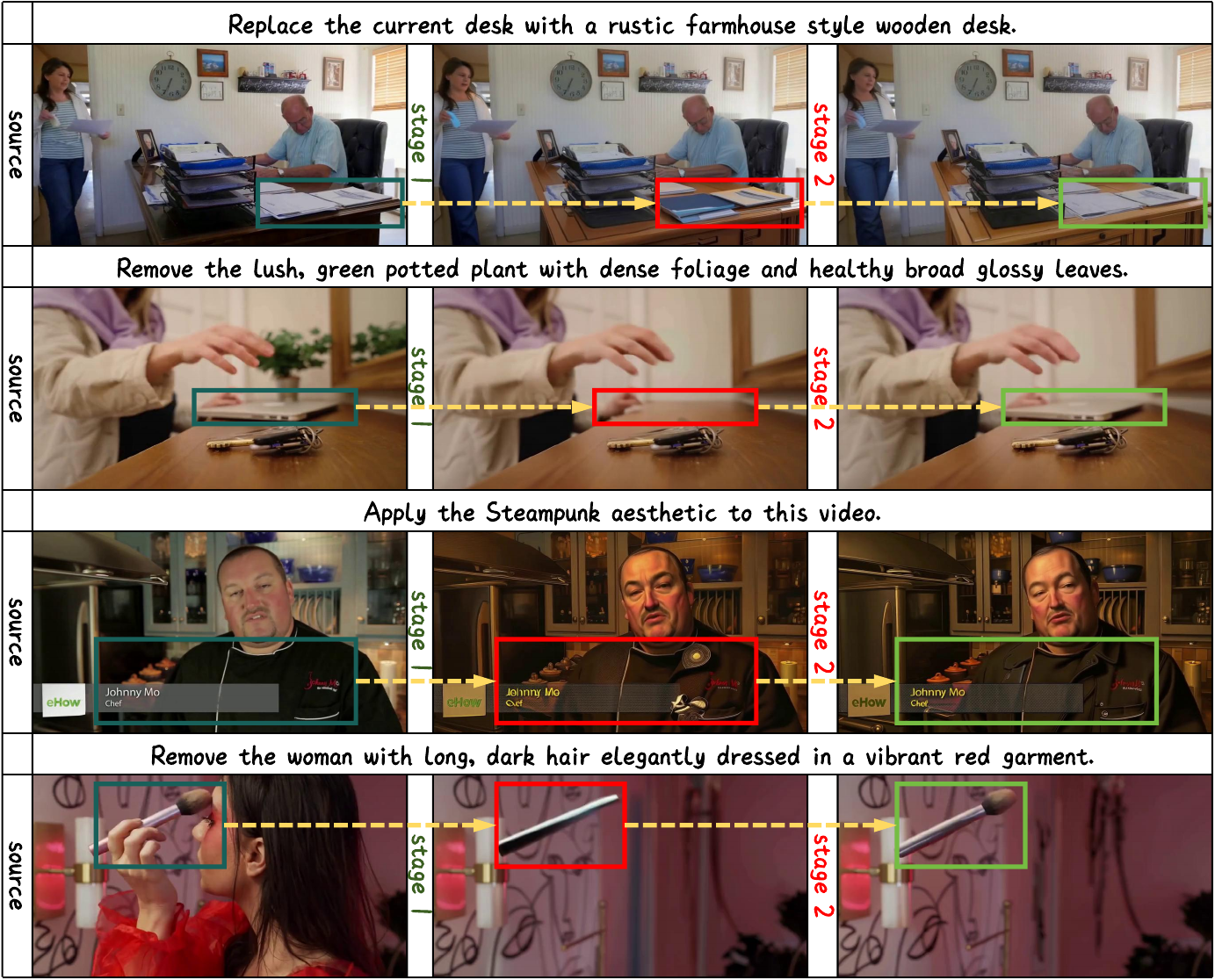}
    \caption{Stage II residual evidence refinement. Comparing the Stage I edited state with its identity-aligned source reference restores source-consistent details in preserved regions.}
    \label{fig:stage2_refinement}
\end{figure}

\begin{table}[!htbp]
\centering
\begin{tabular}{@{}lcccc@{}}
\toprule
Configuration & Overall & G.Style & L.Chg. & Sub. \\
\midrule
\multicolumn{5}{l}{\textit{(a) Source-evidence injection placement}} \\
Head only (1)    & 3.88 & 4.03 & 3.68 & \textbf{4.87} \\
First + last chunks (2) & \textbf{4.03} & \textbf{4.36} & \textbf{3.93} & 4.72 \\
Uniform 3 chunks        & \underline{3.97} & 4.23 & 3.77 & \underline{4.74} \\
All 7 chunks            & 3.95 & \underline{4.29} & \underline{3.78} & 4.62 \\
\midrule
\multicolumn{5}{l}{\textit{(b) Identity supervision ratio}} \\
Identity 0\%  & 3.97 & 4.29 & 3.84 & \textbf{4.75} \\
Identity 5\%  & \underline{4.01} & \textbf{4.39} & \underline{3.87} & 4.71 \\
Identity 10\% & \textbf{4.03} & \underline{4.36} & \textbf{3.93} & \underline{4.72} \\
\midrule
\multicolumn{5}{l}{\textit{(c) MLLM reprompt}} \\
Without reprompt & \underline{3.97} & \underline{4.25} & \underline{3.84} & \underline{4.63} \\
With reprompt    & \textbf{4.03} & \textbf{4.36} & \textbf{3.93} & \textbf{4.72} \\
\midrule
\multicolumn{5}{l}{\textit{(d) Stage II refinement}} \\
Without Stage II & \underline{4.01} & \textbf{4.37} & \underline{3.88} & \underline{4.68} \\
With Stage II    & \textbf{4.03} & \underline{4.36} & \textbf{3.93} & \textbf{4.72} \\
\bottomrule
\end{tabular}
\caption{Ablation studies on OpenVE-Bench using the 2B backbone. Results are ranked within each controlled group.}
\label{tab:ablation}
\end{table}

\subsection{Experimental Analysis}
\label{sec:experimental_analysis}

The main results examine whether separating source-evidence assessment from content synthesis can achieve strong editing with lightweight conditioning. Table~\ref{tab:openve_main} confirms this scale--quality--efficiency balance. With only 37M conditioning parameters, GRNEdit-2B achieves 4.03 Overall and outperforms several larger 5B--14B editors; GRNEdit-8B reaches 4.18 with 45M conditioning parameters, matching the best open-source score. Their per-video inference times are 39\,s and 84\,s, respectively, remaining substantially below most larger baselines. GRNEdit is particularly strong on localization-sensitive tasks, taking the best open-source scores on Background Change and Local Remove.

Fig.~\ref{fig:qualitative} reveals the corresponding control behavior. Across diverse global and local edits, GRNEdit modifies the intended content while preserving source geometry and untargeted subjects. In multi-person scenes, it selects the correct actor and reconstructs removed regions without disturbing the remaining content; competing methods more often under-edit, alter the wrong subject, or disrupt unrelated appearance. This behavior directly supports our \emph{binary evidence perspective}: decoupling edit decisions from semantic synthesis relieves the conditioning branch, reducing parameter overhead while enabling more precise control.

Table~\ref{tab:reco_main} further validates this conclusion on ReCo-Bench, extending evaluation beyond the distribution of our OpenVE-derived training set. Its $S_{\mathrm{VN}}$ and $S_{\mathrm{VQ}}$ further assess motion naturalness, temporal stability, and edit stability. GRNEdit ranks first across all Remove components and reaches 9.08 on Style with the best $S_{\mathrm{VN}}$ and $S_{\mathrm{VQ}}$, demonstrating that GRNEdit’s strengths in edit localization, preservation, and temporal quality transfer robustly to a separate benchmark.

\begin{figure}[!tbp]
    \centering
    \includegraphics[width=\columnwidth]{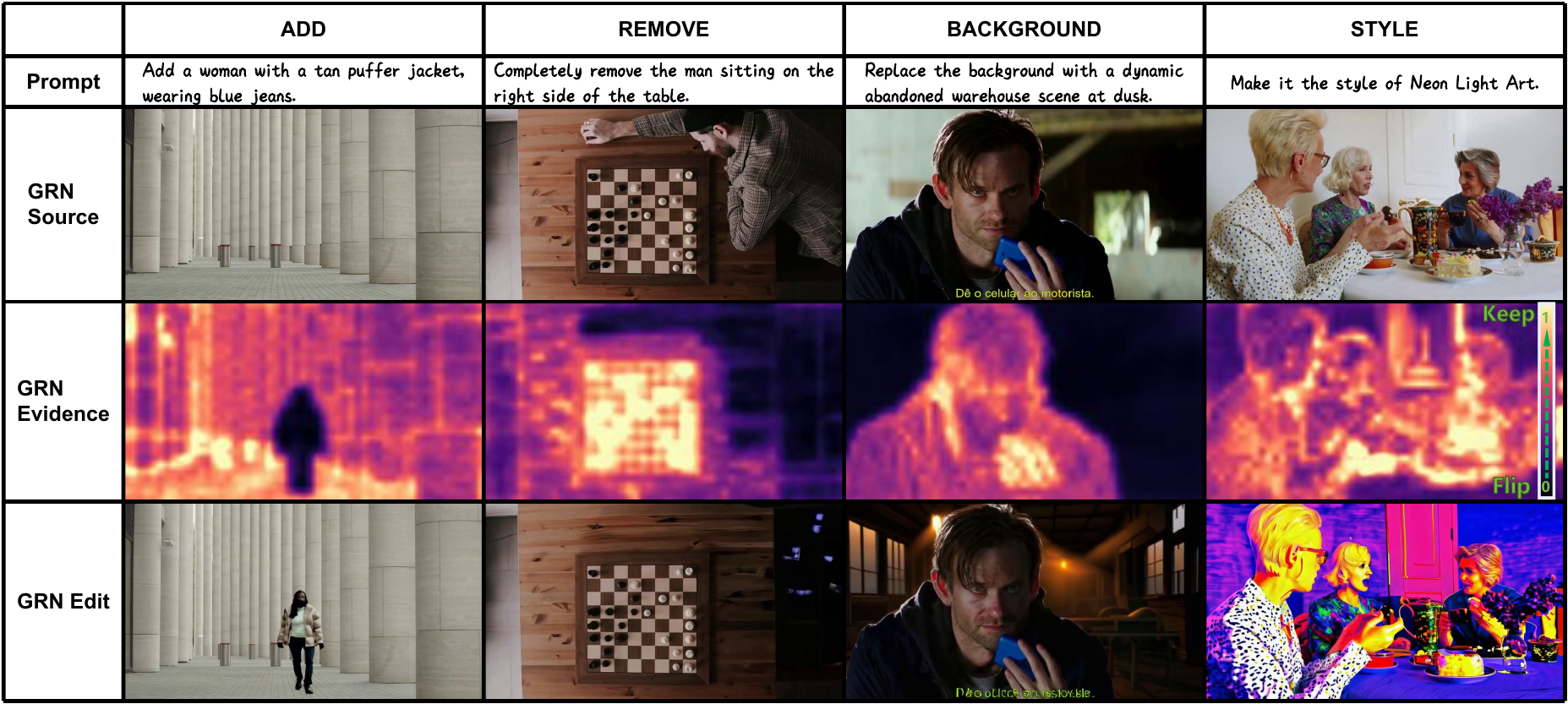}
    \caption{Why call it evidence?
The heatmaps visualize source-induced changes in target-bit margins. Unlike conventional residual representations, source evidence provides measurable, and ordered support for binary editing decisions.}
    \label{fig:evidence_behavior}
\end{figure}

\paragraph{Why Call It Evidence?}
Figure~\ref{fig:evidence_behavior} shows why the source branch functions as evidence rather than a residual representation. We ablate the branch and combine signed margin changes with their RMS magnitude to visualize both the direction and strength of its influence on backbone decisions. The results show that evidence has a clear direction: positive values retain the source bit and negative values flip it. Residual features lack such a shared semantic order and entangle edit location with content. Although attention weights in continuous models may appear probabilistic, they are transient routing coefficients; the representation written back to the backbone comes from value aggregation. In contrast, our support signals act directly on the backbone's native binary decision axis. Empirically, evidence adapts its strength to each task, preserving unedited content while releasing intended edit regions to the backbone. Retained bits receive $6.2\times$ stronger source-aligned responses than changed bits, and strength decreases with source--target disagreement in eight tasks. These directional, spatially selective patterns establish evidence as a functional behavior, not a relabeled residual feature; see Appendix A2.

\begin{figure}[!htbp]
    \centering
    \includegraphics[width=\linewidth]{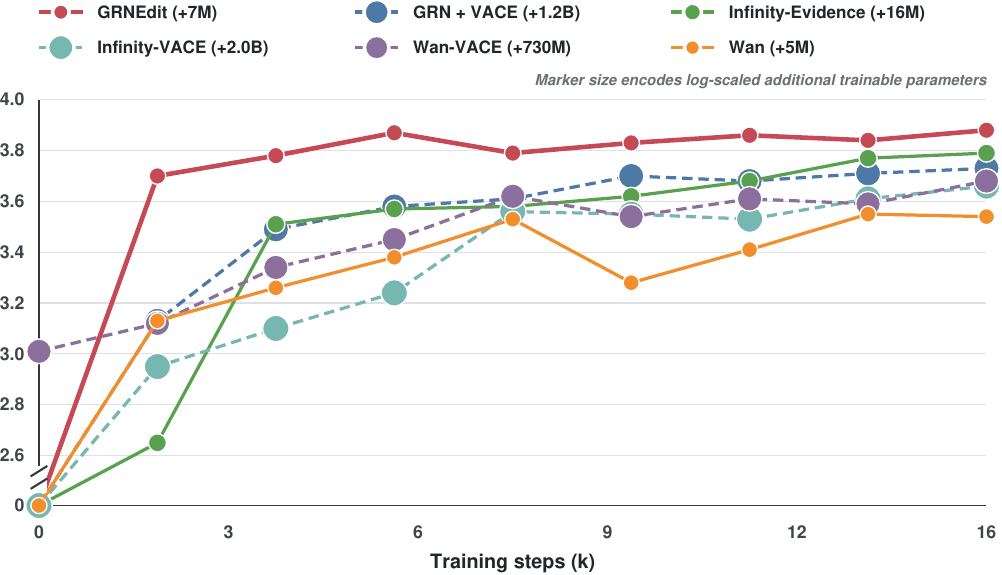}
    \caption{Cross-backbone convergence on OpenVE-Bench. Under matched settings within each backbone, evidence conditioning converges faster than VACE-style conditioning on both GRN and Infinity, while using fewer additional parameters. Marker size denotes additional parameters.}
    \label{fig:convergence_parameter_efficiency}
\end{figure}

\paragraph{Cross-backbone convergence and efficiency.}
Figure~\ref{fig:convergence_parameter_efficiency} compares Evidence and VACE-style conditioning using the closest architecture-compatible implementations; The VACE-style variant copies only the first and last blocks to align with the evidence-based design. On GRN, GRNEdit reaches a competitive 3.70 after only 2K updates and, by 6K, surpasses VACE at 16K with a branch over $100\times$ smaller. Infinity reproduces this ordering on a separate binary backbone, arguing against GRN-specific architecture or pretraining as the sole explanation. Wan provides a complementary control: without binary coordinates, Evidence is approximated by direct residual injection in latent space, where VACE performs better. This representation-dependent reversal, together with faster convergence on both binary backbones, shows that evidence is specifically suited to binary representations, rather than merely a renamed form of conventional conditioning.

\begin{figure}[!htbp]
    \centering
    \includegraphics[width=\linewidth]{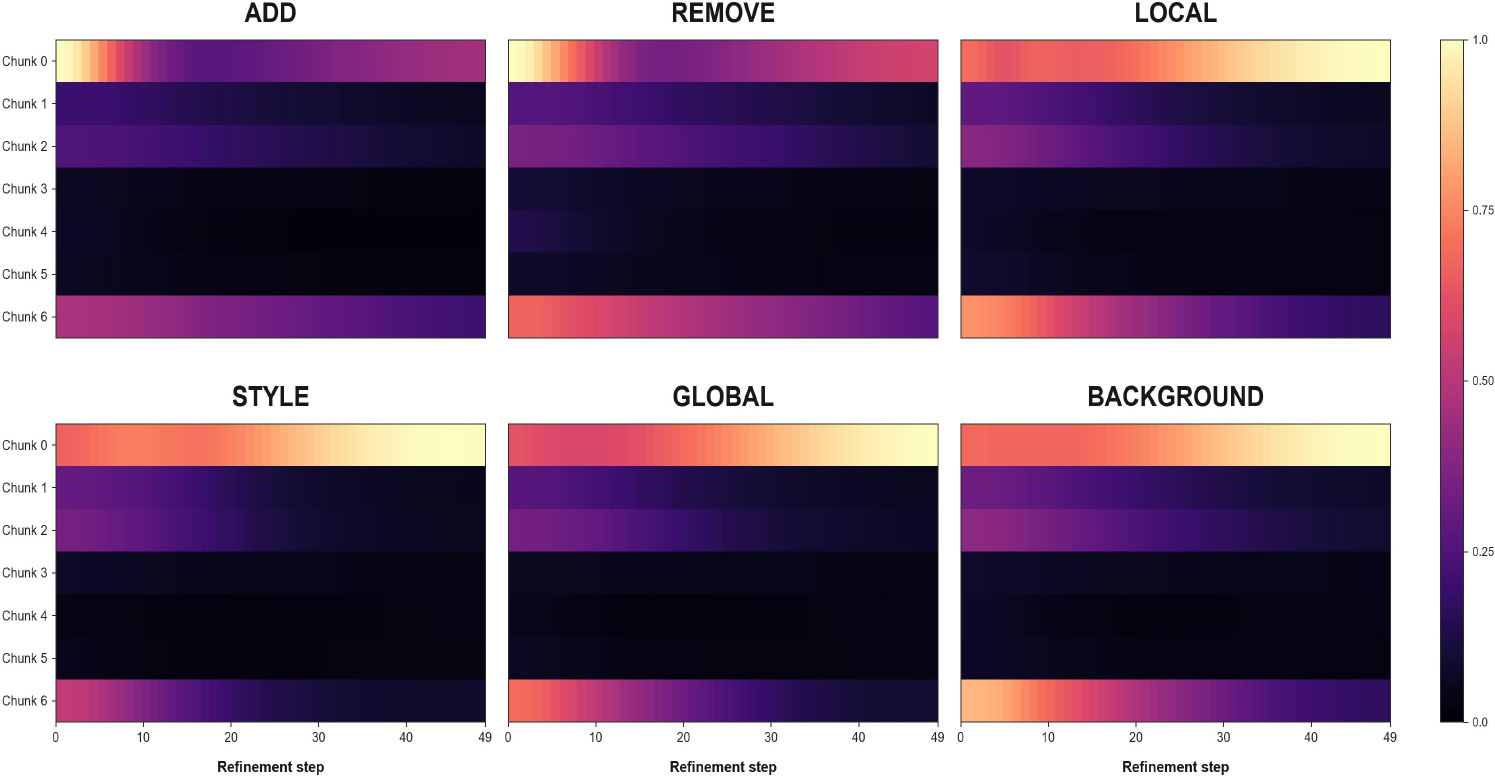}
    \caption{Under a fixed two-injection budget, we retain the first and last chunks, which exhibit the strongest and temporally complementary residual utilization in the fully model.}
    \label{fig:evidence_injection_placement}
\end{figure}

\subsection{Ablation Studies}
\label{sec:ablation}

\paragraph{MLLM reprompt.}
 Removing the reprompt reduces the Overall score from 4.03 to 3.97. Richer scene descriptions help resolve ambiguities in open-ended instructions, while the modest gain confirms that reprompting is a complementary text enhancement rather than the source of performance.

\paragraph{Evidence injection placement.}
Injecting all chunks noticeably slows convergence but brings limited gains in Table \ref{tab:ablation}. We therefore visualize residual utilization relative to the pre-injection hidden states across editing tasks. Figure~\ref{fig:evidence_injection_placement} reveals stronger and temporally complementary responses at the first and last chunks, while intermediate chunks remain weak. Consistently, endpoint injection gives the best Overall, G.Style, and L.Chg. scores (4.03/4.36/3.93), whereas input-only, uniform-three, and all-seven variants reach only 3.88, 3.97, and 3.95 Overall. We thus retain the endpoints for the best accuracy--efficiency trade-off.

\paragraph{Identity-aligned null condition.}
Without identity supervision, the Overall score is 3.97. Using 5\% and 10\% identity samples raises it to 4.01 and 4.03, respectively. The consistent gain shows that aligning the null instruction with source reconstruction provides a clear edit/no-edit anchor and benefits content preservation. Beyond its unexpected improvement in content preservation, this simple objective anchors the edit/no-edit distinction and supplies the source reference connecting Stage~I and Stage~II.

\paragraph{Stage II refinement.}
Stage II adds only 30M parameters. It reuses the identity-aligned path from Stage I as a same-space source reference and learns targeted revisions from the source--generation discrepancy, restoring source-consistent details without suppressing the intended edit. Because VLM-based Overall scoring is  insensitive to pixel-level drift in non-edited regions, the consistency benefit of Stage~II is only weakly reflected in the aggregate score. This benefit is more directly evidenced by a 1.8\,dB PSNR gain (22.0 to 23.8\,dB) over segmentation-identified non-edited regions in OpenVE-Bench's non-global editing tasks and the recovered visual details across replacement and removal tasks in Fig.~\ref{fig:stage2_refinement}.

\section{Conclusion}
\label{sec:conclusion}

We presented GRNEdit, a lightweight two-stage framework that casts general video editing as source-referenced binary decisions, decoupling editing control from content synthesis. Compact source-evidence modules guide target-bit decisions, while identity-aligned null supervision strengthens content preservation and provides an intrinsic reference for residual bit-margin revision. With only 0.6M training pairs and conditioning overhead below 3\% of backbone size, GRNEdit achieves competitive or leading results on OpenVE-Bench and ReCo-Bench, supporting binary evidence as an efficient inductive bias for diverse edits. GRNEdit remains limited in introducing new semantics with little source evidence, such as object addition, and depends on binary backbones. Nevertheless, recent advances, including Infinity and GRN, highlight the growing potential of binary representations as a competitive generative paradigm.

\section*{Acknowledgements}
This work uses the OpenVE-3M, OpenVE-Bench datasets licensed under CC BY-NC 4.0. The ReCo-Bench dataset licensed under CC BY-NC-SA 4.0. The authors confirm that all uses of the above resources are strictly for academic research purposes and not for any commercial application.

\clearpage

\bibliography{ref}

\clearpage
\includepdf[
    pages=-,
    pagecommand={\thispagestyle{empty}},
    fitpaper=true
]{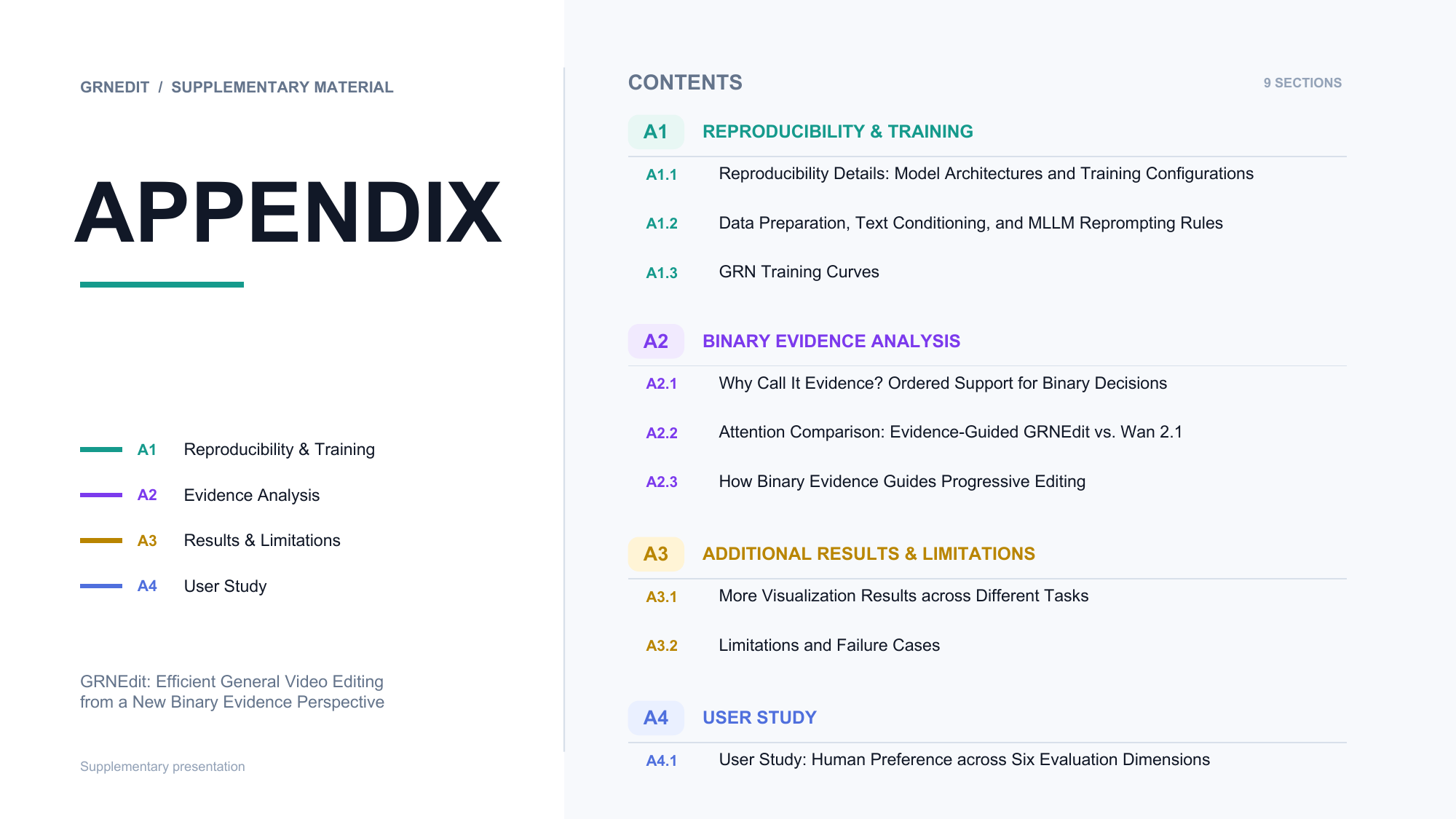}

\end{document}